\documentclass[a4paper]{article}

\usepackage{INTERSPEECH2019}
\usepackage{wrapfig}
\usepackage{url}
\usepackage{setspace}
\usepackage[USenglish]{babel}
\title{Cumulative Adaptation for BLSTM Acoustic Models}
\name{Markus Kitza$^{1}$, Pavel Golik$^{1}$, Ralf Schl\"uter$^{1}$, Hermann Ney$^{1,2}$}

\address{
  $^1$Human Language Technology and Pattern Recognition, Computer Science Department, \\
  RWTH Aachen University, 52074 Aachen, Germany \\
  $^2$AppTek GmbH, 52062 Aachen, Germany \url{http://www.apptek.com/}}
\email{$^1$\{kitza,golik,schlueter,ney\}@cs.rwth-aachen.de}

\begin{document}

\maketitle

{\setstretch{1.2}
\begin{abstract}
	This paper addresses the robust speech recognition problem as an adaptation task.
	Specifically, we investigate the cumulative application of adaptation methods.
	A bidirectional Long Short-Term Memory (BLSTM) based neural network, capable of learning temporal relationships and translation invariant representations, is used for robust acoustic modeling.
	Further, i-vectors were used as an input to the neural network to perform instantaneous speaker and environment adaptation, providing 8\% relative improvement in word error rate on the NIST Hub5 2000 evaluation testset.
	By enhancing the first-pass i-vector based adaptation with a second-pass adaptation using speaker and environment dependent transformations within the network, a further relative improvement of 5\% in word error rate was achieved. We have reevaluated the features used to estimate i-vectors and their normalization to achieve the best performance in a modern large scale automatic speech recognition system.
\end{abstract}
}
\vspace{0.1cm}

\section{Introduction}
The application of deep neural networks to speech recognition has achieved tremendous success due to its superior performance over the traditional hidden Markov model with Gaussian mixture emissions. It has become the dominant acoustic modeling approach for speech recognition, especially for large vocabulary tasks. While it has strong modeling power through multiple layers of nonlinear processing, it is still not immune to many known problems such as the mismatch of training and test data. When tested in mistach conditions, performance degradation can still be expected. To address this problem, many adaptation techniques have been proposed.

Robust speech recognition methods can be classified into two categories: feature-space approaches and model-space approaches.
Compared with model-space approaches, feature-space approaches do not need to modify or retrain the acoustic model.
Instead, various operations can be performed on the acoustic features to improve the noise robustness of the features.
As for the model-space approaches, rather than focusing on the modification of features, the acoustic model parameters are adjusted to match the testing data.

In the category of feature-space approaches popular strategies include using speaker adaptive features~\cite{povey2013}, or augmenting input features with speaker information~\cite{Miao2014c} as well as incorporating auxiliary information such as i-vector and speaker code into the network~\cite{Tan2015,Kundu2016}. Traditionally this also includes feature normalization as the most straightforward strategy to eliminate the training-testing mismatch. This includes strategies like cepstral mean subtraction (CMS)~\cite{Westphal97theuse}, cepstral mean and variance normalization (CMVN)~\cite{molau2003}, and histogram equalization (HEQ)~\cite{hilger2006}.
A further method to increase the robustness against noise is adding a variety of noise samples to clean training data, known as multi-style or multi-condition training~\cite{hirsch2000aurora}.
However, due to the unpredictable nature of real-world noise, it is impossible to account for all noise conditions that may be encountered.

Rather than augmenting the features, the acoustic model parameters can be compensated to match the testing conditions.
A simple example of modifying the models is to re-train the whole speaker independent (SI) deep neural network (DNN) model, or only certain layer(s) of the model on adaptation data~\cite{Liao2013,dong2013}. To avoid over-fitting, regularization such as in \cite{dong2013} is applied. Another approach is to insert and adapt speaker dependent linear layers into the network to transform either input feature~\cite{Neto1995}, top-hidden-layer output~\cite{Li2010}, or hidden layer activations~\cite{Gemello2006}. Finally, the acoustic model can be trained for different conditions separately such as in~\cite{Wu2015a,Delcroix2015,Tan2016}.

This work combines feature-space approaches and model-space approaches and evaluates if they provide complementary improvements in word error rate (WER).
i-vectors~\cite{Dehak2011} are employed and optimized as a feature-space approach and based on our prior work~\cite{kitza18:interspeech}, affine transformations (AT) are used for speaker and environment adaptation.

In this paper, we combine speaker dependent model transforms with i-vectors as an input to the neural network to perform instantaneous speaker and environment adaptation.

To our knowledge i-vectors have not yet been combined with speaker dependent affine transformations within a bidirectional LSTM Network. Therefore, we would like to evaluate how the adaptation performance behaves if they are combined. The effectiveness of adaptation by speaker dependent transformations in regards to the depth of the network is reevaluated in the context of i-vectors. Further, detailed investigations into the structure of the transformations have been done. Also the best methods to train them have been evaluated. We also compare the performance of speaker and environment adaptation.

The remainder of this paper presents our system in detail. Section 2 describes prior work, in section 3 we discuss our implementation of the i-vector adaptation and affine transformation adaptation. Experimental results are analysed in Section 4 followed by a conclusion.

\section{Related work}
The proposed work is builds on our prior work~\cite{kitza18:interspeech}, where we investigated the significance of the position of speaker dependent affine transformations within a bidirectional LSTM Network using a separate transformation for the forward- and backward-direction. It used a similar methodology as presented in \cite{Liu2016} and \cite{Miao2015a}, where affine transformation to adapt an LSTM acoustic model were used. However, here only speaker independent input features were used. Other works in this field include \cite{Zhao2016,Xue2014,Trmal2010,Li2010,Huang2018} and \cite{Gemello2006}, where feedforward neural networks were employed.

I-vectors have been used sucessfully as a sole adaptation method using time-delay neural network (TDNN)~\cite{Peddinti2015} as well as BLSTM acoustic models for automatic speech recognition~\cite{Xiong2018,Kanda2018}.

\section{Adaptation}
In this section, we describe the i-vector estimation process adopted during training and decoding as well as the training procedure for the affine transformations.

\subsection{i-vectors}
In this paper we use a i-vector adapted neural network acoustic model. On each frame we append a i-vector to the 40-dimensional Gammatone Filterbank (GT)~\cite{schluter2007} input of the neural network. Most prior work report that they use Mel-frequency cepstral coefficients (MFCCs)~\cite{Davis1980} to estimate the i-vectors even if other features are used in the acoustic model~\cite{Peddinti2015,Xiong2018,Xiong2016}.
But we noticed that the i-vector adaptation was not sufficiently effective in adapting to test signals when using MFCCs to estimate the i-vectors. Therefore, we compared MFCC to GT features without further processing as well as with concatenated first and second order derivatives and with temporal context and linear discriminant analysis (LDA) for dimension reduction. The results can be seen in Table~\ref{tab:feature}. Using Gammatone features with a context of 9 frames reduced to 60 dimensions with LDA gives us significantly better performance. We did not check if MFCCs would be better if the acoustic model would also be trained on them. 

\begin{table}[th]
	\caption{Comparison of features used for universal background model training. The i-vectors were extraced only from speech frames and have a dimension of 100. Word error rate is given on the full Hub5'00 dataset. The acoustic model is a BLSTM trained on gammatone filterbank features.}
	\label{tab:feature}
	\centering
	\begin{tabular}{l|l|l}
		\textbf{i-vectors}     & \textbf{UBM Features} & \textbf{WER {[}\%{]}} \\
		\hline
		no & ---                   & 14.4                  \\
		\hline
		yes   & MFCC                  & 14.3                  \\\cline{2-3}
		                   & \quad+derivatives      & 14.0                  \\\cline{2-3}
		                   & \quad+context+LDA      & 14.2                  \\\cline{2-3}
		                   & GT                    & 14.2                  \\\cline{2-3}
		                   & \quad+derivatives        & 13.9                  \\\cline{2-3}
		                   & \quad+context+LDA        & \textbf{13.5}                  \\
	\end{tabular}
\end{table}

\subsubsection{i-vector Extraction}
The i-vectors are estimated in the same manner for training and testing datasets. In order to ensure sufficient variety of the i-vectors in the training data, rather than estimating a separate i-vector per speaker, we estimate a single i-vector for each utterance. The i-vectors are estimated only on speech frames. Feature frames which contain silence or noise are discarded prior to the extraction. For the training dataset, the silence frames are classified based on a framewise state alignment obtained from a Hidden markov model with Gaussian mixture emissions system.

For the testing datasets, there are two options. On the one hand, a first pass decode of the audio data using the GMM-HMM system can be used. On the other hand, a two-class Gaussian Mixture Model (GMM) is trained to distinguish between speech and non-speech events to filter out long portions of non-speech data~\cite{magrin2001overview, kingsbury13:cantonese}.

The final part of i-vector extraction is normalization. Rather than using i-vectors that are derived from a total variability model directly, it is typically more feasible to apply some form of normalization first. The basic form is to normalize a given i-vector $v \in \mathbb{R}^D$ with $D \in \mathbb{N}$ to have unit euclidean norm. Another option is to scale $v$ in proportion to the square root of its dimension. The length normalized i-vector $\hat{v}$ is then given by 

\begin{equation*}
	\hat{v} = \frac{v}{||v||_2} \cdot \sqrt{D}.
\end{equation*}

Finally, Radial Gaussianization (RG)~\cite{Lyu2009}, which is used successfully in speaker diarization tasks~\cite{Gracia2011}, can be used for i-vector normalization.

Table~\ref{tab:dim} shows the word error rate given a combination of i-vector dimension and length normalization. It can also be seen, that the adaptation performance dependents significantly on the dimension and normalization.

\begin{table}[th]
	\caption{WERs (in \%) on Hub5'00 for i-vectors of different dimension and normalization based on GT+context+LDA features.}
	\label{tab:dim}
	\centering
	\begin{tabular}{@{}l|l|l|l@{}}
		\textbf{i-vector}               & \multicolumn{3}{c}{\textbf{WER [\%] for Dimension}} \\ 
		\textbf{normalization} & 50   & 100  & 200           \\ \hline
		---                    & 14.9 & 14.3 & 14.8          \\ \hline
		Unity                  & 13.9 & 13.7 & 13.5          \\ \hline
		Square root                   & 14.2 & 13.7 & \textbf{13.3} \\ \hline
		RG                     & 14.0 & 13.4 & 13.4          \\
	\end{tabular}
\end{table}

\subsection{Affine Transformations}
A practical constraint for a large scale speech recognition system is that the system needs to serve many users. Therefore, the user-specific parameters should be kept small. The main goal of this investigation is to develop methods to effectively adapt the speaker independent model using a minimal number of speaker-specific parameters. Two approaches are studied in this work: Adapting existing neural network components and adapting inserted affine transformation between layers.

The affine transformations are realized as additional layers in the neural network. They usually have the same dimension as the preceding layer and the identify function $f(z) = z$ is employed as the activation function for these additional layers. The speaker-specific parameters are given as the weights $W_s$, which are initialized to the unity matrix, and biases $b_s$, which are initialized to $0.0$. These are trained for each speaker separately.

According to the different positions of the linear layers, they are denoted as Linear Input Network (LIN)~\cite{Neto1995}, Linear Hidden Network (LHN)~\cite{Gemello2006} and Linear Output Network (LON)~\cite{Li2010}, where LHN can be inserted to any position between two successive hidden layers. The LIN linearly transforms the observed acoustic features before forwarding them to the speaker independent model, similar to a constrained maximum likelihood linear regression (CMLLR).

When adding a affine transformation to the output layer of the neural network, the transformation is inserted before the softmax function.

A first pass decoding is performed using a speaker-independent model. This is used to generate the targets for the unsupervised adaptation process. The adaptation datasets were split randomly into separate training and cross-validation sets, where 90\% were used for training and 10\% for cross-validation. The cross-validation frame accuracy was also used to control the learning rate decay.

\section{Experimental Results}
The baseline acoustic model was trained on $283$ hours from the Switchboard-1 Release 2 (LDC97S62)~\cite{Godfrey1992} corpus using $40$-dimensional gammatone features without any adaptive feature space transformations, as we did not observe any word error rate reductions with speaker adapted features.
The targets were $9001$ tied states.
The acoustic model consists of seven BLSTM layers for forward and backward direction, each with a size of $500$.
For the training, a dropout~\cite{srivastava2014dropout} probability of $10\%$ is used together with a $L_2$ regularization constant of $0.01$ with an initial learning rate of $0.0005$ that is controlled using the  cross-validation frame accuracy (CVFA) based learning rate decay. This approach divides the learning rate  by $\sqrt{2}$ if the CVFA did not improve. 
For further regularization, gradient noise~\cite{neelakantan2015adding} is added with a variance of $0.3$ and focal loss~\cite{lin2017focal} is used with a factor of $2.0$.
The models have also been pretrained using a layer-wise pretraining algorithm, which gradually builds up the network. 
For the first epoch, only a single layer is used and with each consecutive epoch one additional layers are added until the maximum of seven is reached.

During decoding, we use a $4$-gram language model which was trained on the transcripts of the acoustic training data ($3$M running words) and the transcripts of the Fisher English corpora (LDC2004T19 \& LDC2005T19) with $22$M running words. More details can be found in~\cite{Tuske2015}. The results are reported on the Hub5'00 evaluation data (LDC2002S09) which contains two types of data,  Switchboard (SWBD) -- which is better matched to the training data -- and CallHome English (CHE).

The i-vector estimator was trained on the full 283 hour set of training data: this includes the training of the Gaussian mixture model used for the universal background model (UBM), and the estimation of the total-variability (T) matrix. 

The affine transformation layers are trained using stochastic gradient descent with momentum. In our experience, stochastic gradient descent provides better convergence under a wider set of hyperparameters than more complex algorithms as RMSprop and Nadam. However, the latter show better convergence when the complete acoustic model is trained. The learning rate was set to $10^{-6}$ with a momentum of $0.9$ for all positions. $L_2$-regularization centered on the unity matrix was used with a scale of $0.01$. Beside the \emph{identify} activation function we tried \emph{sigmoid} and \emph{relu} but they consistently underperformed compared to the identify activation function.

\begin{table}[th]
	\caption{WERs (in \%) on Hub5'00 to compare the effectiveness of environment and speaker adaptation at different positions within the acoustic model. The baseline model is a BLSTM without i-vectors.}
	\label{tab:cluster}
	\centering
	\begin{tabular}{l|r|r|r|r|r|r}
		        
		\textbf{Affine}& \multicolumn{6}{c}{\textbf{Adaptation Target}}                              \\
		\textbf{Trans.}          & \multicolumn{3}{c|}{\textbf{Environment}}      & \multicolumn{3}{c}{\textbf{Speaker}} \\
		\textbf{Layer} & \textbf{SWB} & \textbf{CH}   & \multicolumn{1}{l|}{\textbf{Avg.}} & \textbf{SWB} & \textbf{CH}   & \textbf{Avg.} \\ \hline
		---   & 9.7 & 19.1 & 14.4                      & 9.7 & 19.1 & 14.4 \\ \hline
		1     & 9.7 & 18.7 & \textbf{14.2}             & 9.7 & 18.2 & 13.9 \\
		2     & 9.8 & 18.8 & 14.3                      & 9.6 & 18.2 & \textbf{13.8} \\
		3     & 9.8 & 18.7 & 14.3                      & 9.6 & 18.3 & 13.9 \\
		4     & 9.9 & 18.7 & 14.3                      & 9.5 & 18.3 & 13.9 \\
		5     & 9.9 & 19.1 & 14.5                      & 9.6 & 18.5 & 14.1 \\
		6     & 9.8 & 19.1 & 14.5                      & 9.6 & 18.8 & 14.2 \\\hline
		all   & 9.7 & 19.0 & 14.4                      & 9.6 & 18.6 & 14.1 \\
	\end{tabular}
\end{table}

\subsection{Cumulative Adaptation}

Table~\ref{tab:cluster} compares systems without i-vectors but with affine transformations adapted to speakers and environments, using different positions for the transformations. For the environment adaptation, the CallHome and Switchboard subsets were used as environments and for speaker adaptation, each recording was treated as a different speaker. From the table, it is clear, that without i-vectors speaker adaptation outperforms environment adaptation. The results are consistent with~\cite{kitza18:interspeech} in the conclusion, that performing adaptation on single layers at the beginning of a neural network is beneficial compared to adapting later layers or the whole network.

Table~\ref{tab:cluster_ivec} compares systems with i-vectors and affine transformations adapted to speakers and environments. Comparing these to Table~\ref{tab:cluster}, the relative improvements increase. Although the system uses i-vectors internally for adaptation, the additional information provided by the i-vectors is also beneficial for the second pass adaptation. Further, it can be seen that environment adaptation performs better under these circumstances. Using i-vectors, the performance of environment adaptation, where only a single transformation is trained for CallHome and Switchboard respectively, is the same for CallHome with $16.6\%$ and only slightly worse on Switchboard with $8.7\%$ compared to $8.6\%$. Therefore, it is no longer important to train one affine transformations for each speaker, because the transformation can use the information in the i-vector to do the speaker adaptation. Moreover, the best position for environment adaptation is the first layer compared to the second layer for speaker adaptation. Given these circumstances we tried adapting the first and second layer simultaneously, but there were no further improvements to be gained. We also tried adapting the first layer on the environment adaptation set followed by speaker specific adaptation of the second layer. This also gave no additional improvements.  

\begin{table}[th]
	\caption{WERs (in \%) on Hub5'00 to compare the effectiveness of environment and speaker adaptation at different positions within the acoustic model. The baseline model is a BLSTM with i-vectors.}
	\label{tab:cluster_ivec}
	\centering
	\begin{tabular}{l|r|r|r|r|r|r}
		        
		\textbf{Affine}& \multicolumn{6}{c}{\textbf{Adaptation Target}}                              \\
		\textbf{Trans.}          & \multicolumn{3}{c|}{\textbf{Environment}}      & \multicolumn{3}{c}{\textbf{Speaker}} \\
		\textbf{Layer} & \textbf{SWB} & \textbf{CH}   & \multicolumn{1}{l|}{\textbf{Avg.}} & \textbf{SWB} & \textbf{CH}   & \textbf{Avg.} \\ \hline
		---   & 8.9 & 17.7 & 13.3                      & 8.9 & 17.7 & 13.3 \\ \hline
		1     & 8.7 & 16.6 & \textbf{12.7}             & 8.7 & 16.8 & 12.8 \\
		2     & 8.8 & 16.9 & 12.9                      & 8.6 & 16.6 & \textbf{12.6} \\
		3     & 8.9 & 17.0 & 12.9                      & 8.7 & 16.6 & 12.7 \\
		4     & 8.9 & 17.2 & 13.1                      & 8.7 & 16.7 & 12.7 \\
		5     & 8.9 & 17.2 & 13.1                      & 8.8 & 16.9 & 12.8 \\
		6     & 8.9 & 17.2 & 13.1                      & 8.8 & 17.1 & 13.0 \\ \hline
		(1,2) & 8.7 & 16.7 & 12.7                      & 8.6 & 16.7 & 12.7 \\
	\end{tabular}
\end{table}

\subsection{Sequence Training and RNNLMs}
Table~\ref{tab:switchboard} gives detailed word error rates for systems where the cumulative adaptation is used in conjuction with a lattice-based version of state-level minimum Bayes risk (SMBR) training as well as recurrent neural network language models (RNNLM)~\cite{beck19:interspeech}. Similar to the second pass adaptation also SMBR training provides larger relative improvements if i-vectors are used in the baseline.

\begin{table}[h]
	\caption{Detailed WERs (in \%) on Hub5'00 comparing the influence of i-vectors, SMBR, and environment adaptation with affine transformation (AT). WERs are given for count based language models and RNNLM.}
	\label{tab:switchboard}
	\centering
						        
	\begin{tabular}{llllrrr}
		\textbf{}                        & \textbf{}                           & \textbf{}                          & \multicolumn{1}{l|}{\textbf{}}   & \multicolumn{3}{c}{\textbf{Hub5'00}}                                                                     \\
		\multicolumn{1}{l|}{\textbf{AT}} & \multicolumn{1}{l|}{\textbf{i-vec.}} & \multicolumn{1}{l|}{\textbf{SMBR}} & \multicolumn{1}{l|}{\textbf{LM}} & \multicolumn{1}{l|}{\textbf{SWBD}} & \multicolumn{1}{l|}{\textbf{CH}} & \multicolumn{1}{l}{\textbf{Avg.}} \\ \hline
		\multicolumn{1}{l|}{no}          & \multicolumn{1}{l|}{no}             & \multicolumn{1}{l|}{no}            & \multicolumn{1}{l|}{4-gram}      & \multicolumn{1}{r|}{9.7}           & \multicolumn{1}{r|}{19.1}        & 14.4                              \\ \cline{3-7} 
		\multicolumn{1}{l|}{}            & \multicolumn{1}{l|}{}               & \multicolumn{1}{l|}{yes}           & \multicolumn{1}{l|}{4-gram}      & \multicolumn{1}{r|}{9.6}           & \multicolumn{1}{r|}{18.3}        & 13.9                              \\ \cline{4-7} 
		\multicolumn{1}{l|}{}            & \multicolumn{1}{l|}{}               & \multicolumn{1}{l|}{}              & \multicolumn{1}{l|}{LSTM}         & \multicolumn{1}{r|}{7.7}           & \multicolumn{1}{r|}{15.3}        & 11.7                              \\ \cline{2-7} 
		\multicolumn{1}{l|}{}            & \multicolumn{1}{l|}{yes}            & \multicolumn{1}{l|}{no}            & \multicolumn{1}{l|}{4-gram}      & \multicolumn{1}{r|}{8.9}           & \multicolumn{1}{r|}{17.7}        & 13.3                              \\ \cline{3-7} 
		\multicolumn{1}{l|}{}            & \multicolumn{1}{l|}{}               & \multicolumn{1}{l|}{yes}           & \multicolumn{1}{l|}{4-gram}      & \multicolumn{1}{r|}{8.3}           & \multicolumn{1}{r|}{16.7}        & 12.5                              \\ \cline{4-7} 
		\multicolumn{1}{l|}{}            & \multicolumn{1}{l|}{}               & \multicolumn{1}{l|}{}              & \multicolumn{1}{l|}{LSTM}         & \multicolumn{1}{r|}{6.7}           & \multicolumn{1}{r|}{14.7}        & 10.7                              \\ \cline{1-4} \cline{4-7} 
		\multicolumn{1}{l|}{yes}         & \multicolumn{1}{l|}{yes}               & \multicolumn{1}{l|}{yes}              & \multicolumn{1}{l|}{4-gram}      & \multicolumn{1}{r|}{8.1}           & \multicolumn{1}{r|}{15.4}        & 11.8                              \\ \cline{4-7} 
		\multicolumn{1}{l|}{}            & \multicolumn{1}{l|}{}               & \multicolumn{1}{l|}{}              & \multicolumn{1}{l|}{LSTM}         & \multicolumn{1}{r|}{6.7}           & \multicolumn{1}{r|}{13.5}        & 10.2                              \\ \hline
																		 &                                     &                                    &                                  & \multicolumn{1}{l}{}               & \multicolumn{1}{l}{}             & \multicolumn{1}{l}{}             
		\end{tabular}
\end{table}

\section{Conclusion}
Using a combination of i-vectors and environment dependent unsupverised second pass training of affine transformatons, we were able to show that the cumulative application of these adaptation methods gives significantly larger improvements than any method on its own.
The choice of features and normalization for i-vector estimation was shown to have a large influence on their adaptation performance.
Also, we have shown, that environment adaptation and speaker adaptation perform best at different locations within the network.

Our best single system achieves a word error rate of $10.2\%$ on the Hub5'00 evaluation corpus when trained only on 283 hours  of training data. To our knowledge this is state of the art for a recognition system not based on system combination.

\section{Acknowledgements}
This work has received funding from the European Research Council (ERC) under the European Union's Horizon 2020 research and innovation programme (grant agreement No 694537, project "SEQCLAS" and Marie Sk\l{}odowska-Curie grant agreement No 644283, project "LISTEN") and from a Google Focused Award. 
The work reflects only the authors' views and none of the funding parties is responsible for any use that may be made of the information it contains. 

\bibliographystyle{IEEEtran}

\bibliography{mybib}

\begin{thebibliography}{10}
\providecommand{\url}[1]{#1}
\csname url@samestyle\endcsname
\providecommand{\newblock}{\relax}
\providecommand{\bibinfo}[2]{#2}
\providecommand{\BIBentrySTDinterwordspacing}{\spaceskip=0pt\relax}
\providecommand{\BIBentryALTinterwordstretchfactor}{4}
\providecommand{\BIBentryALTinterwordspacing}{\spaceskip=\fontdimen2\font plus
\BIBentryALTinterwordstretchfactor\fontdimen3\font minus
  \fontdimen4\font\relax}
\providecommand{\BIBforeignlanguage}[2]{{%
\expandafter\ifx\csname l@#1\endcsname\relax
\typeout{** WARNING: IEEEtran.bst: No hyphenation pattern has been}%
\typeout{** loaded for the language `#1'. Using the pattern for}%
\typeout{** the default language instead.}%
\else
\language=\csname l@#1\endcsname
\fi
#2}}
\providecommand{\BIBdecl}{\relax}
\BIBdecl

\bibitem{povey2013}
P.~S. Rath, D.~Povey, K.~Vesel{\'{y}}, and J.~{\v{C}}ernock{\'{y}}, ``Improved
  feature processing for deep neural networks,'' in \emph{Proceedings of
  Interspeech 2013}, no.~8, Lyon, France, 2013, pp. 109--113.

\bibitem{Miao2014c}
Y.~Miao, L.~Jiang, H.~Zhang, and F.~Metze, ``{Improvements to speaker adaptive
  training of deep neural networks},'' in \emph{2014 IEEE Spoken Language
  Technology Workshop (SLT)}.\hskip 1em plus 0.5em minus 0.4em\relax South Lake
  Tahoe, NV, USA: IEEE, Dec. 2014, pp. 165--170.

\bibitem{Tan2015}
Y.~Qian, T.~Tan, D.~Yu, and Y.~Zhang, ``{Integrated adaptation with
  multi-factor joint-learning for far-field speech recognition},'' in
  \emph{2016 IEEE International Conference on Acoustics, Speech and Signal
  Processing (ICASSP)}, Shanghai, China, March 2016, pp. 5770--5774.

\bibitem{Kundu2016}
S.~Kundu, G.~Mantena, Y.~Qian, T.~Tan, M.~Delcroix, and K.~C. Sim, ``{Joint
  acoustic factor learning for robust deep neural network based automatic
  speech recognition},'' in \emph{2016 IEEE International Conference on
  Acoustics, Speech and Signal Processing (ICASSP)}, Shanghai, China, March
  2016, pp. 5025--5029.

\bibitem{Westphal97theuse}
M.~Westphal, ``The use of cepstral means in conversational speech
  recognition,'' in \emph{In Proceedings of the European Conference on Speech
  Communication and Technology (Eurospeech}, Rhodes, Greece, September 1997,
  pp. 1143--1146.

\bibitem{molau2003}
S.~Molau, F.~Hilger, and H.~Ney, ``Feature space normalization in adverse
  acoustic conditions,'' in \emph{IEEE International Conference on Acoustics,
  Speech, and Signal Processing}, vol.~1, Hong Kong, China, Apr. 2003, pp.
  656--659.

\bibitem{hilger2006}
F.~{Hilger} and H.~{Ney}, ``Quantile based histogram equalization for noise
  robust large vocabulary speech recognition,'' vol.~14, no.~3, Toulouse,
  France, May 2006, pp. 845--854.

\bibitem{hirsch2000aurora}
H.-G. Hirsch and D.~Pearce, ``The aurora experimental framework for the
  performance evaluation of speech recognition systems under noisy
  conditions,'' in \emph{ASR2000-Automatic Speech Recognition: Challenges for
  the new Millenium ISCA Tutorial and Research Workshop (ITRW)}, 2000.

\bibitem{Liao2013}
H.~Liao, ``Speaker adaptation of context dependent deep neural networks,'' in
  \emph{2013 IEEE International Conference on Acoustics, Speech and Signal
  Processing}, Vancouver, BC, Canada, May 2013, pp. 7947--7951.

\bibitem{dong2013}
D.~Yu, K.~Yao, H.~Su, G.~Li, and F.~Seide, ``{KL-Divergence Regularized Deep
  Neural Network Adaptation For Improved Large Vocabulary Speech
  Recognition},'' in \emph{IEEE International Conference on Acoustics, Speech
  and Signal Processing}, Vancouver, BC, Canada, May 2013, pp. 7893--7897.

\bibitem{Neto1995}
J.~Neto, L.~Almeida, M.~Hochberg, C.~Martins, L.~Nunes, S.~Renals, and
  T.~Robinson, ``{Speaker-Adaptation for Hybrid HMM-ANN Continuous Speech
  Recognition System},'' no. September, Madrid, Spain, September 1995, pp.
  2171--2174.

\bibitem{Li2010}
B.~Li and K.~C. Sim, ``{Comparison of Discriminative Input and Output
  Transformations for Speaker Adaptation in the Hybrid NN/HMM Systems},'' in
  \emph{Interspeech}, Makuhari, Chiba, Japan, Sept. 2010.

\bibitem{Gemello2006}
R.~Gemello, F.~Mana, S.~Scanzio, P.~Laface, and R.~{De Mori}, ``{Adaptation of
  Hybrid ANN/HMM Models Using Linear Hidden Transformations and Conservative
  Training},'' in \emph{IEEE International Conference on Acoustics Speed and
  Signal Processing Proceedings}, Toulouse, France, May 2006, pp.
  I--1189--I--1192.

\bibitem{Wu2015a}
C.~Wu and M.~J. Gales, ``{Multi-basis adaptive neural network for rapid
  adaptation in speech recognition},'' in \emph{2015 IEEE International
  Conference on Acoustics, Speech and Signal Processing (ICASSP)}.\hskip 1em
  plus 0.5em minus 0.4em\relax Brisbane, QLD, Australia: IEEE, Apr. 2015, pp.
  4315--4319.

\bibitem{Delcroix2015}
M.~Delcroix, K.~Kinoshita, T.~Hori, and T.~Nakatani, ``{Context adaptive deep
  neural networks for fast acoustic model adaptation},'' in \emph{2016 IEEE
  International Conference on Acoustics, Speech and Signal Processing
  (ICASSP)}, Shanghai, China, March 2016, pp. 4535--4539.

\bibitem{Tan2016}
T.~Tan, Y.~Qian, and K.~Yu, ``{Cluster Adaptive Training for Deep Neural
  Network Based Acoustic Model},'' \emph{IEEE/ACM Transactions on Audio,
  Speech, and Language Processing}, vol.~24, no.~3, pp. 459--468, March 2016.

\bibitem{Dehak2011}
N.~Dehak, P.~J. Kenny, R.~Dehak, P.~Dumouchel, and P.~Ouellet, ``{Front-end
  factor analysis for speaker verification},'' vol.~19, no.~4, Prague, Czech
  Republic, May 2011, pp. 788--798.

\bibitem{kitza18:interspeech}
M.~Kitza, R.~Schl\"uter, and H.~Ney, ``Comparison of blstm-layer-specific
  affine transformationsfor speaker adaptation,'' in \emph{Interspeech},
  Hyderabad, India, Sep. 2018, pp. 877--881.

\bibitem{Liu2016}
C.~Liu, Y.~Wang, K.~Kumar, and Y.~Gong, ``{Investigations on speaker adaptation
  of LSTM RNN models for speech recognition},'' in \emph{IEEE International
  Conference on Acoustics, Speech and Signal Processing (ICASSP)}, Shanghai,
  China, March 2016, pp. 5020--5024.

\bibitem{Miao2015a}
Y.~Miao and F.~Metze, ``{On speaker adaptation of long short-term memory
  recurrent neural networks},'' in \emph{Interspeech}, Dresden, Germany, Sept.
  2015.

\bibitem{Zhao2016}
Y.~Zhao, J.~Li, and Y.~Gong, ``{Low-rank plus diagonal adaptation for deep
  neural networks},'' in \emph{IEEE International Conference on Acoustics,
  Speech and Signal Processing (ICASSP)}.\hskip 1em plus 0.5em minus
  0.4em\relax Shanghai, China: IEEE, March 2016, pp. 5005--5009.

\bibitem{Xue2014}
J.~Xue, J.~Li, D.~Yu, M.~Seltzer, and Y.~Gong, ``{Singular value decomposition
  based low-footprint speaker adaptation and personalization for deep neural
  network},'' in \emph{IEEE International Conference on Acoustics, Speech and
  Signal Processing (ICASSP)}, Florence, Italy, May 2014, pp. 6359--6363.

\bibitem{Trmal2010}
J.~Trmal, J.~Zelinka, and L.~M{\"u}ller, ``Adaptation of a feedforward
  artificial neural network using a linear transform,'' in \emph{Text, Speech
  and Dialogue}, P.~Sojka, A.~Hor{\'a}k, I.~Kope{\v{c}}ek, and K.~Pala,
  Eds.\hskip 1em plus 0.5em minus 0.4em\relax Berlin, Heidelberg: Springer
  Berlin Heidelberg, 2010, pp. 423--430.

\bibitem{Huang2018}
\BIBentryALTinterwordspacing
Z.~Huang, H.~Lu, M.~Lei, and Z.~Yan, ``{Linear networks based speaker
  adaptation for speech synthesis},'' March 2018. [Online]. Available:
  \url{http://arxiv.org/abs/1803.02445}
\BIBentrySTDinterwordspacing

\bibitem{Peddinti2015}
\BIBentryALTinterwordspacing
V.~Peddinti, G.~Chen, V.~Manohar, T.~Ko, D.~Povey, and S.~Khudanpur, ``{JHU
  ASpIRE system: Robust LVCSR with TDNNS, iVector adaptation and RNN-LMS},'' in
  \emph{2015 IEEE Workshop on Automatic Speech Recognition and Understanding
  (ASRU)}.\hskip 1em plus 0.5em minus 0.4em\relax IEEE, dec 2015, pp. 539--546.
  [Online]. Available: \url{http://ieeexplore.ieee.org/document/7404842/}
\BIBentrySTDinterwordspacing

\bibitem{Xiong2018}
W.~{Xiong}, L.~{Wu}, F.~{Alleva}, J.~{Droppo}, X.~{Huang}, and A.~{Stolcke},
  ``The microsoft 2017 conversational speech recognition system,'' in
  \emph{2018 IEEE International Conference on Acoustics, Speech and Signal
  Processing (ICASSP)}, Calgary, Alberta, Canada, April 2018, pp. 5934--5938.

\bibitem{Kanda2018}
N.~Kanda, R.~Ikeshita, S.~Horiguchi, Y.~Fujita, K.~Nagamatsu, X.~Wang,
  V.~Manohar, N.~Yalta, M.~Maciejewski, S.-J. Chen, A.~Shanmugam~Subramanian,
  R.~Li, Z.~Wang, J.~Naradowsky, L.~Paola Garcia-Perera, and G.~Sell, ``The
  hitachi/jhu chime-5 system: Advances in speech recognition for everyday home
  environments using multiple microphone arrays,'' 09 2018, pp. 6--10.

\bibitem{schluter2007}
R.~{Schl\"uter}, I.~{Bezrukov}, H.~{Wagner}, and H.~{Ney}, ``Gammatone features
  and feature combination for large vocabulary speech recognition,'' in
  \emph{2007 IEEE International Conference on Acoustics, Speech and Signal
  Processing - ICASSP '07}, vol.~4, Honolulu, Hawaii, USA, April 2007, pp.
  IV--649--IV--652.

\bibitem{Davis1980}
S.~{Davis} and P.~{Mermelstein}, ``Comparison of parametric representations for
  monosyllabic word recognition in continuously spoken sentences,'' \emph{IEEE
  Transactions on Acoustics, Speech, and Signal Processing}, vol.~28, no.~4,
  pp. 357--366, August 1980.

\bibitem{Xiong2016}
\BIBentryALTinterwordspacing
W.~Xiong, J.~Droppo, X.~Huang, F.~Seide, M.~Seltzer, A.~Stolcke, D.~Yu, and
  G.~Zweig, ``{The Microsoft 2016 Conversational Speech Recognition System},''
  \emph{Arxiv}, no.~Lm, 2016. [Online]. Available:
  \url{http://arxiv.org/abs/1609.03528}
\BIBentrySTDinterwordspacing

\bibitem{magrin2001overview}
I.~Magrin-Chagnolleau, G.~Gravier, and R.~Blouet, ``Overview of the 2000-2001
  {ELISA} consortium research activities,'' in \emph{2001: A Speaker
  Odyssey-The Speaker Recognition Workshop}, 2001.

\bibitem{kingsbury13:cantonese}
B.~Kingsbury, J.~Cui, X.~Cui, M.~Gales, K.~Knill, J.~Mamou, L.~Mangu,
  D.~Nolden, M.~Picheny, B.~Ramabhadran, R.~Schl{\"u}ter, A.~Sethy, and
  P.~Woodland, ``A high-performance cantonese keyword search system,'' in
  \emph{IEEE International Conference on Acoustics, Speech, and Signal
  Processing}, Vancouver, BC, Canada, May 2013, pp. 8277--8281.

\bibitem{Lyu2009}
\BIBentryALTinterwordspacing
S.~Lyu and E.~P. Simoncelli, ``Nonlinear extraction of independent components
  of natural images using radial gaussianization,'' \emph{Neural Computation},
  vol.~21, no.~6, pp. 1485--1519, 2009, pMID: 19191599. [Online]. Available:
  \url{https://doi.org/10.1162/neco.2009.04-08-773}
\BIBentrySTDinterwordspacing

\bibitem{Gracia2011}
D.~Garcia-Romero and C.~Espy-Wilson, ``Analysis of i-vector length
  normalization in speaker recognition systems.'' Florence, Italy, August 2011,
  pp. 249--252.

\bibitem{Godfrey1992}
J.~Godfrey, E.~Holliman, and J.~McDaniel, ``{SWITCHBOARD: telephone speech
  corpus for research and development},'' in \emph{ICASSP-92: 1992 IEEE
  International Conference on Acoustics, Speech, and Signal Processing}.\hskip
  1em plus 0.5em minus 0.4em\relax San Francisco, CA, USA: IEEE, March 1992,
  pp. 517--520.

\bibitem{srivastava2014dropout}
N.~Srivastava, G.~Hinton, A.~Krizhevsky, I.~Sutskever, and R.~Salakhutdinov,
  ``Dropout: a simple way to prevent neural networks from overfitting,''
  \emph{The Journal of Machine Learning Research}, vol.~15, no.~1, pp.
  1929--1958, 2014.

\bibitem{neelakantan2015adding}
A.~Neelakantan, L.~Vilnis, Q.~V. Le, I.~Sutskever, L.~Kaiser, K.~Kurach, and
  J.~Martens, ``Adding gradient noise improves learning for very deep
  networks,'' \emph{arXiv preprint arXiv:1511.06807}, 2015.

\bibitem{lin2017focal}
T.-Y. Lin, P.~Goyal, R.~Girshick, K.~He, and P.~Doll{\'a}r, ``Focal loss for
  dense object detection,'' \emph{arXiv preprint arXiv:1708.02002}, 2017.

\bibitem{Tuske2015}
Z.~Tuske, P.~Golik, R.~Schl\"uter, and H.~Ney, ``Speaker adaptive joint
  training of gaussian mixture models and bottleneck features,'' in \emph{2015
  {IEEE} Workshop on Automatic Speech Recognition and Understanding
  ({ASRU})}.\hskip 1em plus 0.5em minus 0.4em\relax Scottsdale, AZ, USA:
  {IEEE}, Dec. 2015.

\bibitem{beck19:interspeech}
E.~Beck, W.~Zhou, R.~Schl\"uter, and H.~Ney, ``{LSTM} language models for
  {LSCVR} in first-pass decoding and lattice-rescoring,'' in \emph{submitted to
  Interspeech}, 2019.

\end{thebibliography}

\end{document}